\documentclass{article}

\usepackage[utf8]{inputenc}
\usepackage[T1]{fontenc}

\usepackage{amsmath,amsfonts,amssymb,times,graphicx,natbib,algorithm,algorithmic,hyperref}

\usepackage{microtype}
\usepackage{graphicx}
\usepackage{subfig}
\usepackage{booktabs}
\usepackage{amsmath}
\usepackage{mathtools}
\usepackage{amsfonts}
\usepackage{ifthen}

\DeclareMathOperator*{\argmax}{arg\,max}
\DeclarePairedDelimiter{\norm}{\lVert}{\rVert}

\usepackage[accepted]{whi2020}

\icmltitlerunning{Visualizing Transfer Learning}

\begin{document}

\twocolumn[

\icmltitle{Visualizing Transfer Learning}

\icmlsetsymbol{equal}{*}

\begin{icmlauthorlist}
\icmlauthor{Róbert Szabó}{elte}
\icmlauthor{Dániel Katona}{bme}
\icmlauthor{Márton Csillag}{elte}
\icmlauthor{Adrián Csiszárik}{renyi,elte}
\icmlauthor{Dániel Varga}{renyi}
\end{icmlauthorlist}

\icmlaffiliation{elte}{Eötvös Loránd University, Budapest, Hungary}
\icmlaffiliation{bme}{Budapest University of Technology and Economics, Budapest, Hungary}
\icmlaffiliation{renyi}{Alfréd Rényi Institute of Mathematics, Budapest, Hungary}

\icmlcorrespondingauthor{Dániel Varga}{daniel@renyi.hu}

\icmlkeywords{feature visualization, transfer learning}

\vskip 0.3in
]

\printAffiliationsAndNotice{}

\begin{abstract}
We provide visualizations of individual neurons of a deep image recognition network during the temporal process of transfer learning. These visualizations qualitatively demonstrate various novel properties of the transfer learning process regarding the speed and characteristics of adaptation, neuron reuse, spatial scale of the represented image features, and behavior of transfer learning to small data. We publish the large-scale dataset that we have created for the purposes of this analysis.
\end{abstract}

\section{Introduction}

Deep neural networks are still commonly conceptualized as black boxes, despite all the recent progress made in interpretability and feature visualization \citep{buhrmester2019analysis, olah2020zoom, hohman2019summit, rathore2019topoact, bau2017network, Selvaraju_2019}.

The current work is following in the footsteps of the Clarity research programme \citep{olah2017feature, olah2018the, carter2019activation, olah2020zoom}, both in the techniques employed, and in the qualitative flavour of the research: creating images of neurons, and trying to identify interesting patterns.

Our main focus is using feature visualization to get a better understanding of what happens during transfer learning, both by comparing neurons before-and-after transfer learning, and by observing what happens during the transfer learning process. We also present a channel visualization technique employing a learned prior \citep{synthesizing2016} utilizing the StyleGAN2 generator \citep{karras2020analyzing}.

Another output of the current work is a large-scale visualization of the transfer learning behavior of an InceptionV1 network. This mapping of the InceptionV1 network with its 57 convolutional layers and 7280 channels on four datasets resulted in approximately 30 000 visualization images. The produced dataset is presented in a browsable form at \url{https://bit.ly/visualizing-transfer-learning}.

\begin{figure}[t]
    \centering
    \includegraphics[width=\columnwidth]{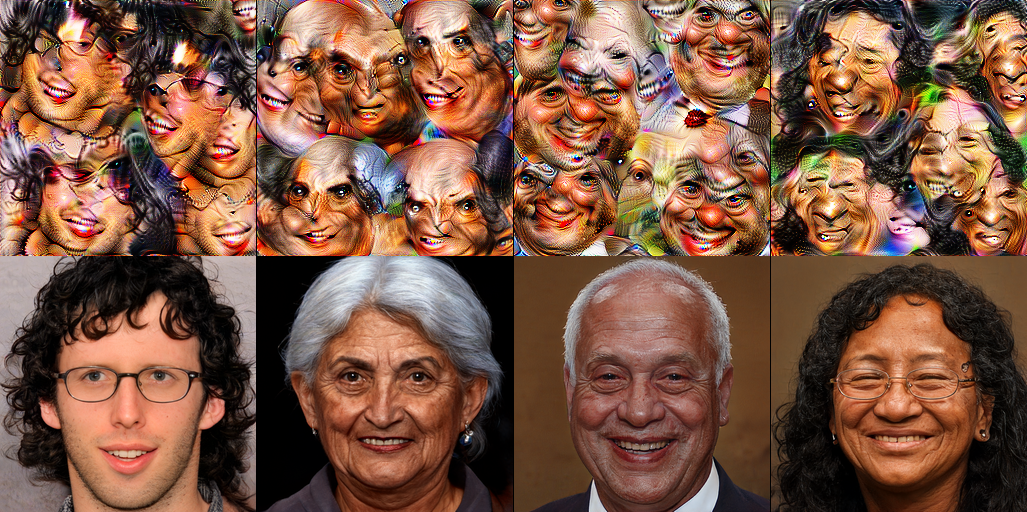}
    \caption{Visualizations for layer $\texttt{Mixed\_5c\_Branch\_3\_b\_1x1}$ channels. Columns correspond to channels, top row shows Lucid visual, bottom row shows StyleGAN2-based visual. Note the strong correspondence of facial features between the two kinds of visualizations.}
    \label{fig:lucid_stylegan}
\end{figure}

\section{Feature visualization via activation maximization}

Gradient based methods of feature visualization strive to maximize the aggregated activation of a network layer, channel, or single neuron by computing the activation's gradient with respect to the input image, and doing gradient ascent \citep{zeiler2014, mahendran2015, simonyan2013deep, mordvintsev2015inceptionism, olah2017feature}. To achieve good results, some image parametrizations or priors must be added that guide the optimization process to an output interpretable to human observers.

Usually the goal is to add priors that bring the least amount of their own biases, but a particularly interesting exception is the use of generative models: feeding the output of an independently trained generator to the inspected model, and doing regularized gradient ascent in the latent space of the generator \citep{synthesizing2016}. We employ this technique on a CelebA classifier's top convolutional layer, using the StyleGAN2 generator \citep{karras2020analyzing}. In effect, we can solve the highly nonlinear activation maximization of face recognition neurons, constrained to the manifold of face images.

More precisely, with an $F: \mathbb{R}^{w_1 \times h_1 \times 3} \to \mathbb{R}^{m}$ recognition network, and a $G:  \mathbb{R}^{d} \to \mathbb{R}^{w_2 \times h_2 \times 3}$ synthesis network (where $w_1, h_1$ and $w_2, h_2$ are the spatial dimensions of the images for the two models, respectively), we solve the following soft-constrained optimization task using SGD:
$$ w^* = \argmax_{w \in \mathbb{R}^{d}} F_{z}(downscale(G(w))) - \lambda \norm{w - \hat{w}}^2, $$
where $F_{z}$ is the average-pooled activation function of channel $z$, $\hat{w}$ is the center of gravity of StyleGAN2's intermediate $W$ space \citep{karras2020analyzing}, $\lambda$ is a multiplier hyperparameter tuning the trade-off between more realistic/typical faces and faces activating neuron $z$. The optimization is started from $\hat{w}$. The visualization obtained is $G(w^*)$.

It is important to note that gradient-based feature visualization (even when combined with e.g. the Lucid framework's diversity feature \citep{olah2017feature}) only presents a facet of the functionality of a channel or neuron, and higher layer neurons are probably always multi-faceted \citep{szegedy2013intriguing, olah2020zoom}. We will limit ourselves to observations that do not assume that an apparent functionality of a channel is its \emph{only} functionality.

\section{Setup}

Unless otherwise noted, all our experiments use the Lucid framework for visualizing convolutional channels via activation maximization \citep{olah2017feature}. We use Lucid's 2D FFT image representation with decorrelation. Lucid has the capability to optimize individual neural activations within a channel, and its authors had success with optimizing the spatially central neuron of each channel, but as we achieved better results on our tasks when optimizing the average-pooled activations of channels, this is what we use in all our presented visualizations.

The network that we analyze in all of our experiments is an InceptionV1 \citep{szegedy2015going}, with two dense layers at the top (these are: dense(1024, Relu), dense(number\_of\_classes, activation\_func), where activation\_func is softmax for Flowers17 and Animals, and coordinate-wise sigmoid for CelebA. Accordingly, the loss functions were categorical cross-entropy and mean coordinate-wise binary cross-entropy respectively.

The InceptionV1 network is a standard choice in the feature visualization community. After all the progress in classification performance, this network still appears to be the best option if the goal is easy to interpret gradient-based feature visualization. (We are not aware of any explanation of this phenomenon.) We however deviate from \citep{olah2017feature} in using a batch-normalized InceptionV1 variant. An unfortunate minor side effect of this choice is the lack of direct correspondence between our neurons and the Activation Atlas \citep{carter2019activation}.

We employ three datasets for the transfer learning task:
The CelebFaces Attributes Dataset (CelebA) \citep{liu2018large} is a large-scale face attributes dataset with more than 200K face images, each annotated with 40 binary attributes. The Flowers17 dataset \citep{nilsback2006visual} contains 17 flower categories with 80 images for each class. The Animals dataset contains 31 dog and cat breed categories, 200 image of each.

In each case, we start from network weights trained on ImageNet.

No layers were frozen during transfer learning. The networks were trained for 200 epochs with the Adam optimizer with learning rate 0.001 and batch size of 32.

Validation accuracies were 0.94 for Animals, 0.99 for Flowers17, mean binary accuracy was 0.90 for CelebA. Feature visualizations were created based on these networks.

Visualization of the temporal process of transfer learning used a different setup. An InceptionV1 network was trained on the CelebA dataset for 3000 iterations with batch size 10. The low batch size was chosen to show a finer detail about the first few iterations that already result in large changes in the visualized features. The top only had a single dense layer.

A technical detail regarding the StyleGAN2 synthesis network is that it also has noise variables as input. In each gradient ascent step these are sampled, leading to nondeterministic results, as we will present in Figure~\ref{fig:repeated_runs}.

\section{Discussion}

\begin{figure*}[t]
    \centering
    \caption{Visualization of transfer learning from ImageNet to CelebA at different layer depths. Each row corresponds to a single layer, odd columns correspond to layer's first 4 channels pre-transfer, even columns are the same neurons post-transfer. The selected layers are every 7th layer ending with the deepest layer, namely $\texttt{Conv2d\_2b\_1x1}$, $\texttt{Mixed\_3b\_Branch\_3\_b\_1x1}$, $\texttt{Mixed\_4b\_Branch\_1\_a\_1x1}$, $\texttt{Mixed\_4c\_Branch\_2\_a\_1x1}$, $\texttt{Mixed\_4d\_Branch\_0\_a\_1x1}$, $\texttt{Mixed\_4e\_Branch\_1\_b\_3x3}$, $\texttt{Mixed\_4f\_Branch\_2\_b\_3x3}$, $\texttt{Mixed\_5b\_Branch\_3\_b\_1x1}$. (The periodicity of the InceptionV1 layers is 6.) Layers are deeper from top to down.}
    \vspace{1em}
    \includegraphics[width=\textwidth]{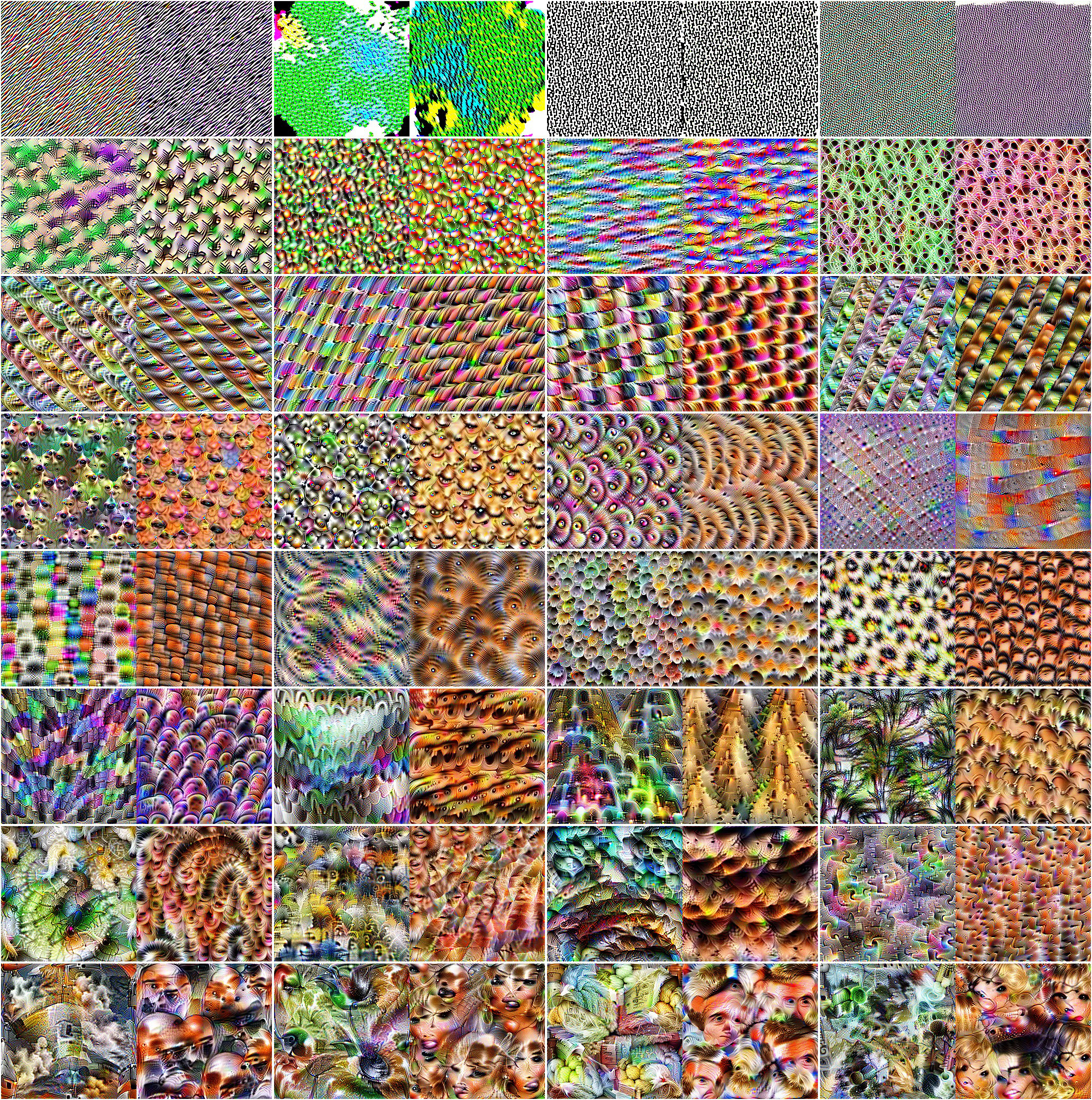}
    \label{fig:montage}
\end{figure*}

In this section we present qualitative observations we made while inspecting our visualizations. We highlight visual evidence for each, but as there is a thin line between highlighting and cherry-picking, we encourage the reader to browse the complete set of visualizations at \url{https://bit.ly/visualizing-transfer-learning} to verify the claims or make their own observations. Figure~\ref{fig:montage} also presents a completely unbiased sample of channels. 
Forming and validating quantitative hypotheses based on the observations is an important further step that we intend to make in follow-up work.

\begin{figure*}[ht]
    \centering
    \caption{Visualization of transfer learning (from ImageNet to CelebA) at different training iterations. Columns show distinct channels, rows show the following iterations, respectively: 0, 10, 20, 30, 60, 150, 1000, 3000. Note that for this experiment the batch size is 10 to visualize a finer grained detail of the transfer learning process.}%
    \vspace{1em}
    \subfloat[$\texttt{4c\_2\_b\_3x3}$]{{\includegraphics[width=2.5cm]{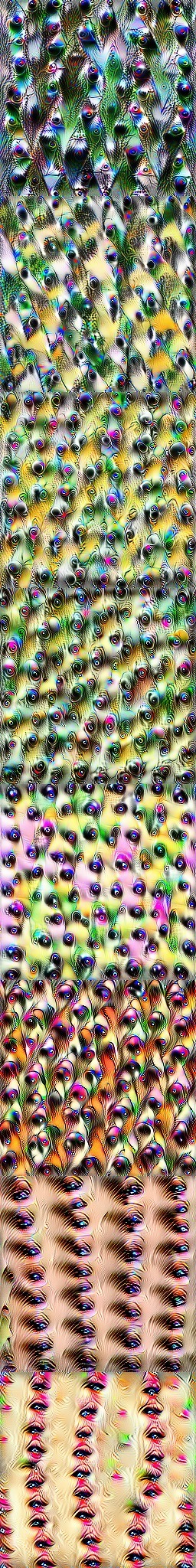} }}%
    \subfloat[$\texttt{4d\_3\_b\_1x1}$]{{\includegraphics[width=2.5cm]{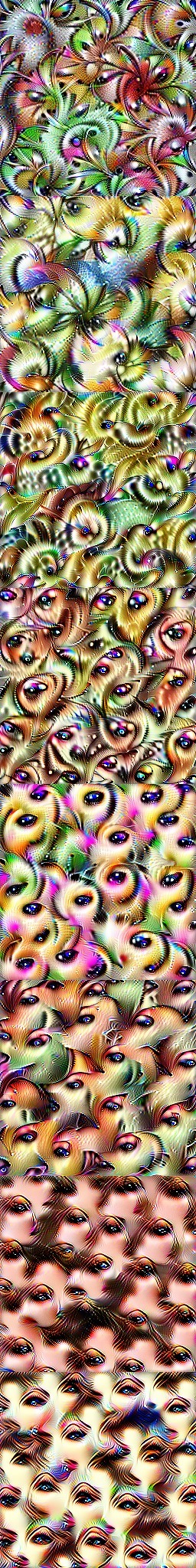} }}%
    \subfloat[$\texttt{4e\_2\_a\_1x1}$]{{\includegraphics[width=2.5cm]{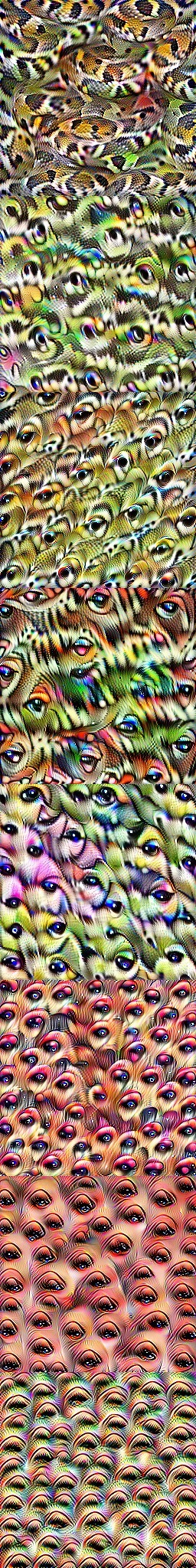} }}%
    \subfloat[$\texttt{4f\_1\_a\_1x1}$]{{\includegraphics[width=2.5cm]{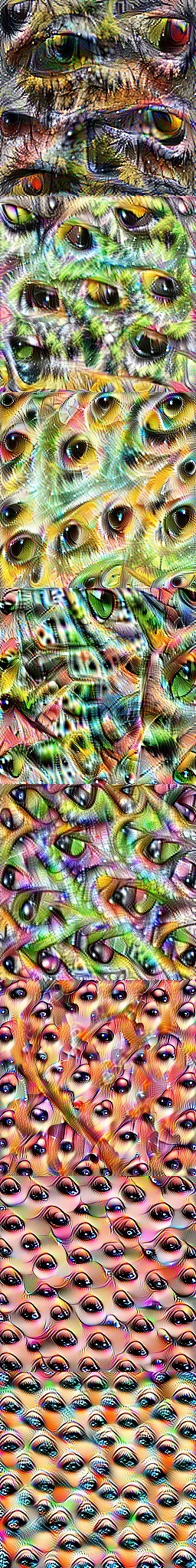} }}%
    \subfloat[$\texttt{5b\_3\_b\_1x1}$]{{\includegraphics[width=2.5cm]{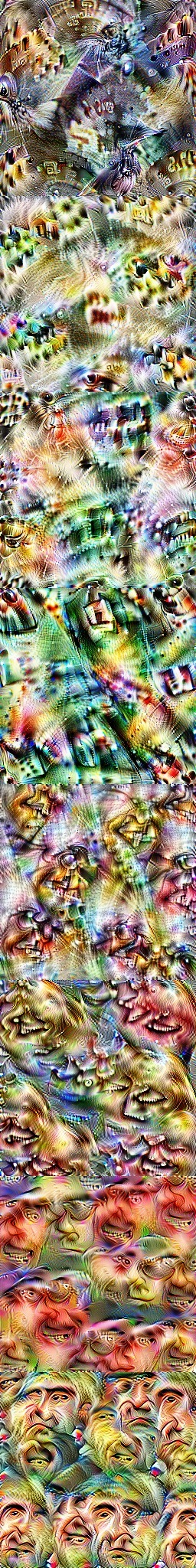} }}%
    \subfloat[$\texttt{5c\_0\_a\_1x1}$]{{\includegraphics[width=2.5cm]{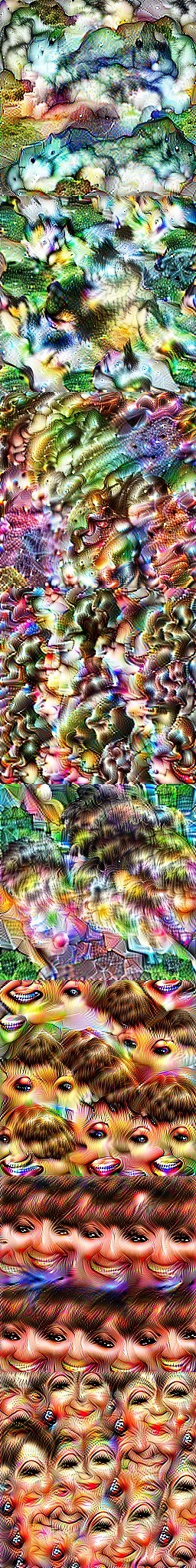} }}%
    \label{fig:transfer_iterations}%
\end{figure*}

\paragraph{Adaptation happens early in the transfer learning process.} While monitoring the progress of transfer learning, the feature visualizations show a surprisingly early emergence of features from the target domain. The visualizations of Figure~\ref{fig:transfer_iterations} show that in the case of adapting e.g. from ImageNet to CelebA, several characteristic CelebA features appear after training on only 100-600 images from the target domain (2nd to 4th row in Figure~\ref{fig:transfer_iterations}).
Note that while ImageNet data contains images of human faces, these are on a very different spatial scale compared to CelebA data. We provide some before-transfer feature visualizations for comparison on Figure~\ref{fig:montage}. They rarely show highly similar patterns at the higher layers, and this fact strongly suggests that the face features that emerge early are acquired during transfer learning rather than already latently present in the ImageNet model.

\begin{figure}[t]
    \centering
    \includegraphics[width=\columnwidth]{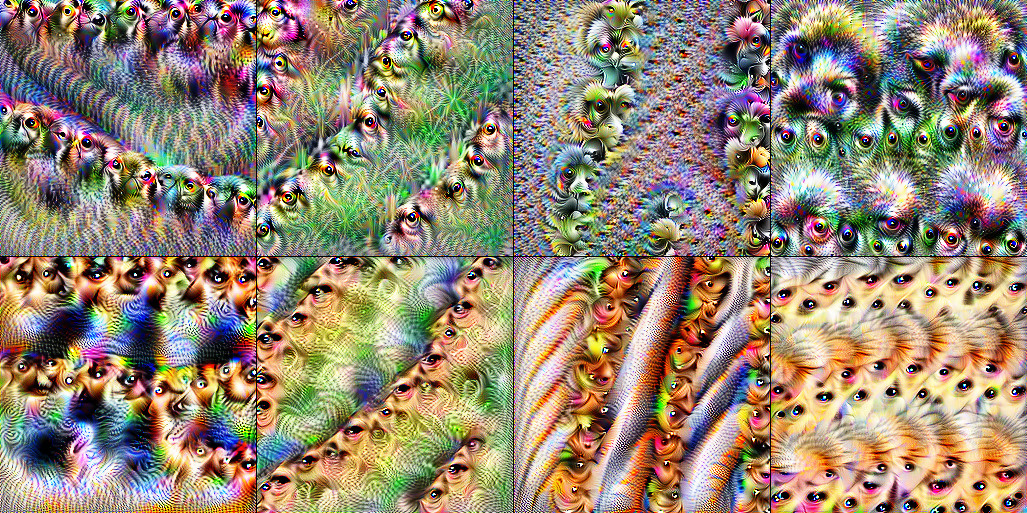}
    \caption{Top row shows pre transfer, bottom row shows post transfer (from ImageNet to CelebA) neuron visualizations. Layer: $\texttt{Mixed\_4c\_Branch\_2\_b\_3x3}$.}
    \label{fig:high_level_transfer}
\end{figure}

\paragraph{Even middle layer detectors can be reused.}
Contrasting with, but not contradicting the previous point, it does happen that CelebA facial feature detectors are direct descendants of ImageNet feature detectors. For example, ImageNet contains many animal faces at the right scale, and detectors for these face parts are often subtly adjusted for the target domain, as seen on Figure~\ref{fig:high_level_transfer}.

\begin{figure}[t]
    \centering
    \includegraphics[width=\columnwidth]{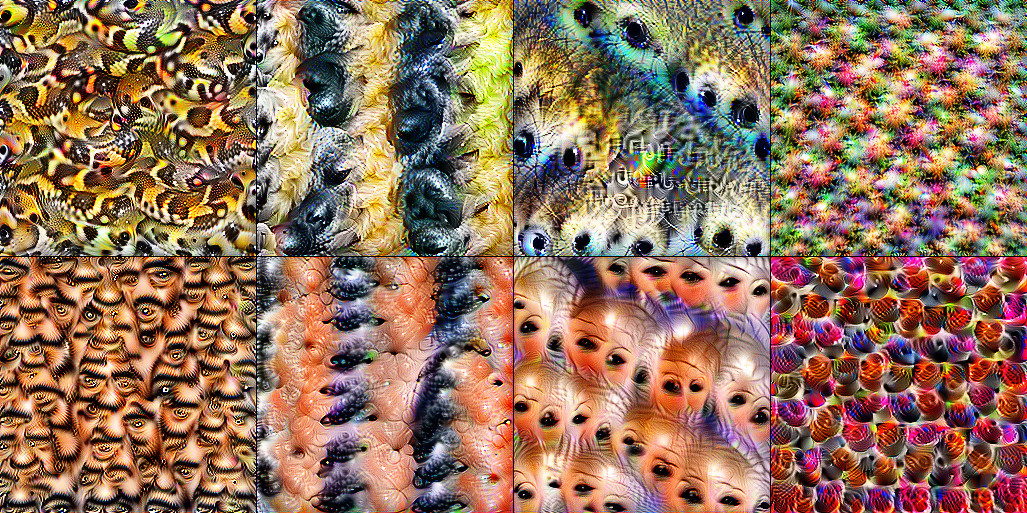}
    \caption{Channels from a single layer. Each channel has a fundamental spatial scale, and it is stable during transfer learning. Layer:  $\texttt{Mixed\_4e\_Branch\_2\_b\_3x3}$.}
    \label{fig:receptive_field}
\end{figure}

\paragraph{Each channel has a fundamental spatial scale, and it is stable during transfer.} Our channel-wise optimized visualizations have a distinct organic repeating pattern, and the period can change from 4 pixels to circa 120 pixels. (The InceptionV1 input size is $224 \times 224$.) Each layer has a predominant period, and this value increases with the depth of the layer, as expected. But this period can still vary significantly within a single layer, and this demonstrates that neurons have an effective receptive field size which is not necessarily the same as their convolutional receptive field size. Remarkably, this effective receptive field size tends to be stable during transfer learning. Figure~\ref{fig:receptive_field} and Figure~\ref{fig:transfer_iterations} demonstrate the phenomenon.

\paragraph{If the dataset is small, the channels solve the same tasks repeatedly.} When comparing the results of the large CelebA dataset with the small Animals and Flowers17 datasets, it is apparent that in the latter cases, the higher level neurons do not manage to find a diverse enough set of features, and this manifests itself in high redundancy across channels: very often it is impossible to distinguish the visuals of two channels. In principle, this could be an artifact of the optimization process used for feature visualization, but its prevalence and consistency suggests otherwise. The paper only presents visualizations on CelebA, we refer the reader to our webpage \url{https://bit.ly/visualizing-transfer-learning} for the Animals and Flowers17 collection.

\paragraph{The lower layers rarely adapt besides adjusting color space.} Below the $\texttt{Mixed\_3c}$ layer, the predominant form of adaptation is adjusting the color scheme the channel is most interested in, without altering the pattern. Moving further across layers, first subtle, then less subtle changes appear when comparing the pre- and post-transfer visuals. See the first two rows of Figure~\ref{fig:montage} for visualizations at lower depths.

\paragraph{Image-scale structures can emerge.} A peculiar property of InceptionV1 feature visualizations is that the middle layers are the most interpretable to the human eye. We report that this is an artifact of the thematically diverse ImageNet dataset, and with our datasets the top layers are the most interpretable. In particular, the top convolutional layers of our CelebA network are apparently sensitive to complete face images, and the visualizations reproduce much of the complexity of human faces, see Figure~\ref{fig:lucid_stylegan} top row visualizing some top convolutional layer neurons.

\subsection{Feature Visualization with a Generative Prior}

Our generator based visualizations are mostly believable as real human faces, although they have some of the characteristics of caricatures, exaggerating facial features and overrepresenting unusual shapes and forms when utilizing smaller $\lambda$ values.

\begin{figure}[t]
    \centering
    \includegraphics[width=\columnwidth]{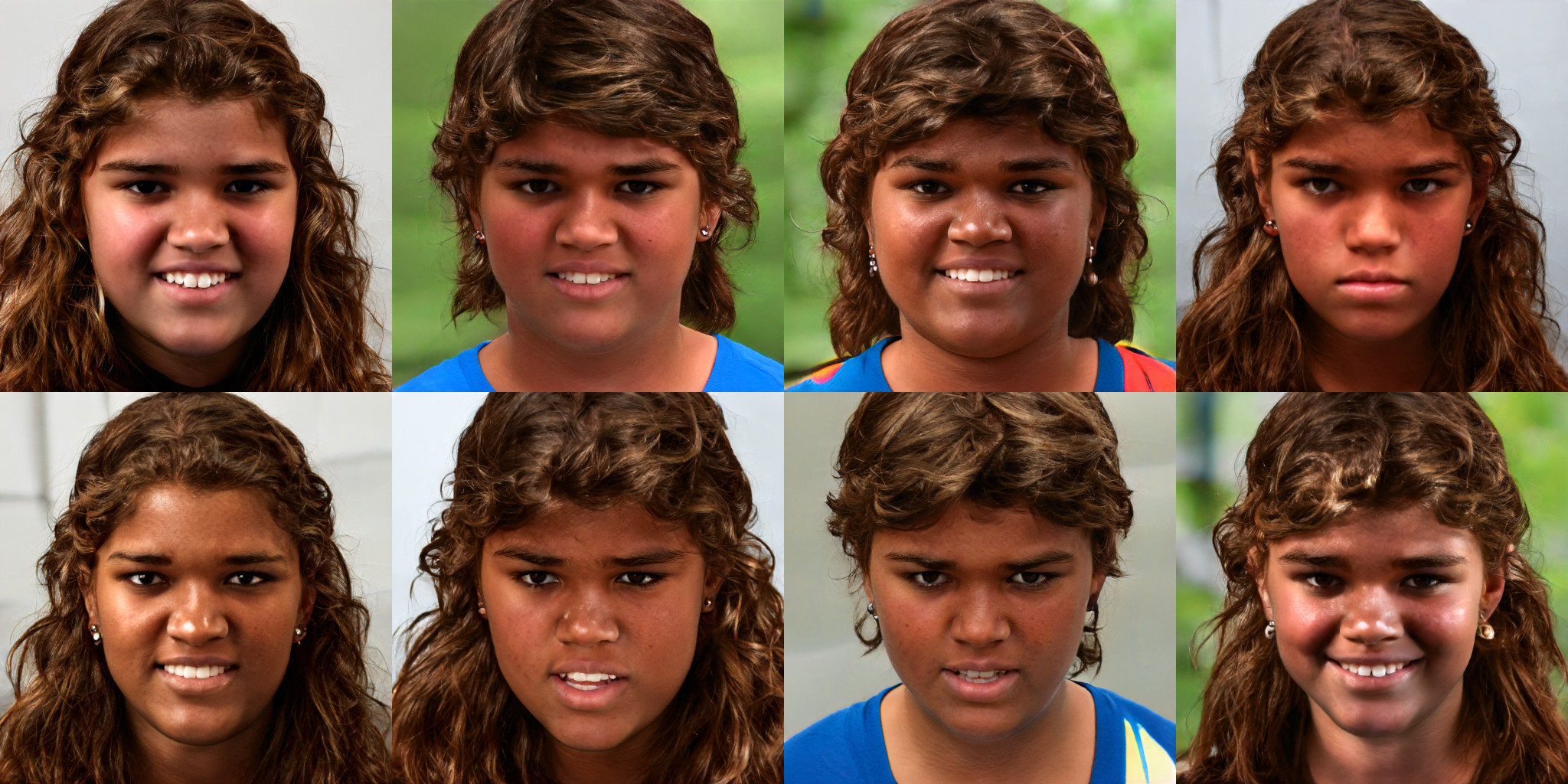}
    \caption{Repeated runs result in similar but not identical images. Utilized layer for visualization: $\texttt{Mixed\_5c\_Branch\_3\_b\_1x1}$.}
    \label{fig:repeated_runs}
\end{figure}

Repeated runs tend to converge to similar but not identical latent points and generated images, see Figure~\ref{fig:repeated_runs}. Note that the only source of randomness during optimization is the noise input of the synthesis network, so the technique gives only a limited view into the true diversity of local optima.

Importantly, comparing the Lucid image and the corresponding StyleGAN2 image as done on Figure~\ref{fig:lucid_stylegan}, we can see that they reinforce and refine each other's message. Reinforce in the sense that organic face-like patterns of Lucid and the photorealistic StyleGAN2 images often seem to picture the same (nonexistent) person, or at least they present unique facial features appearing on both images. Refine in the sense that some features that are quite robustly manifested even across many repeated runs, can turn out to be an artifact of the optimization process, for example the gender of the person. See the accompanying webpage \url{https://bit.ly/visualizing-transfer-learning} for more examples.

\begin{figure}[t]
    \centering
    \includegraphics[width=\columnwidth]{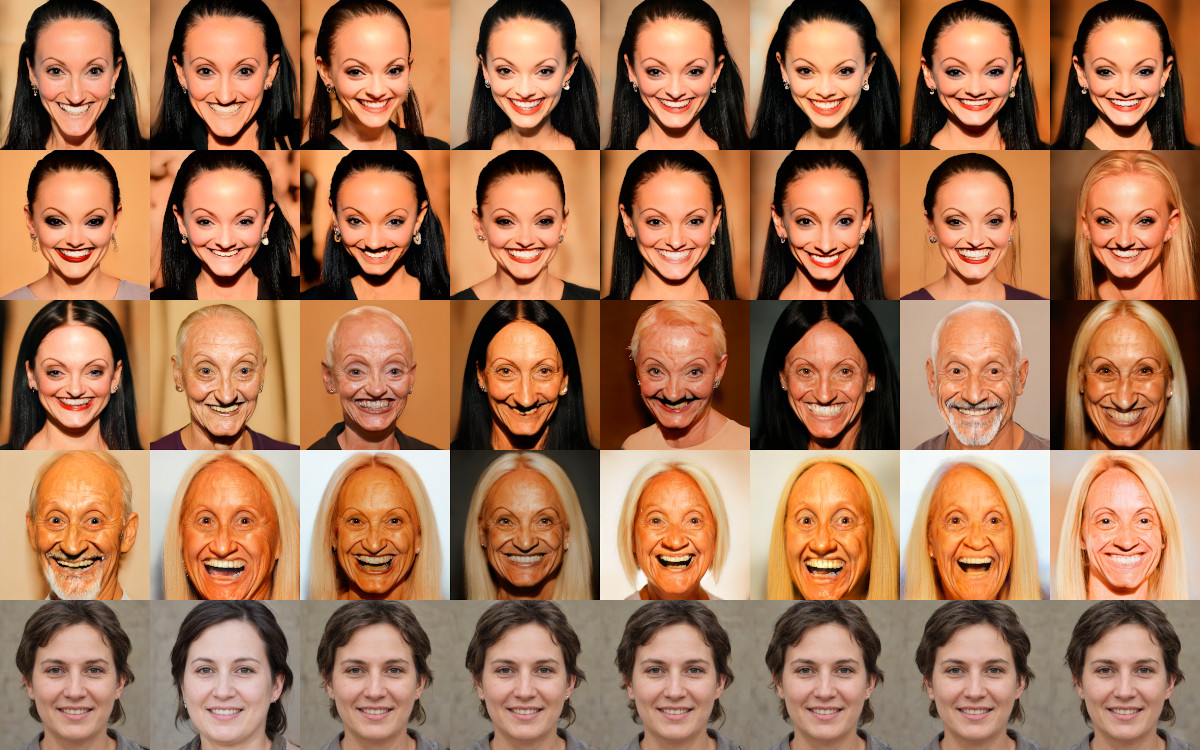}
    \caption{Neural weight ablation: filter weights of a single channel of the top convolutional layer were ranked by magnitude. The first $k$ weights were negated, and the ablated channel was visualized with the generative prior. $k$ running from 0 to 39, presented in row major order on a grid.}
    \label{fig:ablation}
\end{figure}

In a crude preliminary approximation of \textit{circuit editing} \citep{olah2020zoom} presented in Figure~\ref{fig:ablation}, we have modified the weights of a single top convolutional layer neuron, and visualized the result with the generative prior. The modification strategy was picking the top $k$ filter weights by magnitude and negate them. Most of the weight modifications do not meaningfully affect the visualization outcome, but some affect them significantly. In the last elements of the array we can see that at some point, damaging the filter makes it impossible for the optimizer to move away from the $\hat{w}$ prior.

\section{Conclusion}

In this work we provided visualizations of individual neurons of a deep image recognition network during the temporal process of transfer learning. These visualizations qualitatively demonstrate various novel properties of the transfer learning process regarding the speed and characteristics of adaptation, neuron reuse, spatial scale of the represented image features, and behavior of transfer learning to small data.

Even though we tried to limit our exposure to this effect by promoting observations that show an unambiguous signal, and can in principle be formalized and quantitatively verified, the main risk of the qualitative approach we pursue is that the human brain is prone to making up stories where information is ambiguous. Hence, utilizing human visual pattern matching must be paired with some quantitative follow-up analysis when the goal is to make claims about neural networks. This is our ongoing work.

\section*{Acknowledgements}

We thank Dániel Feles for designing the website for the project.

This work was supported by the European Union, co-financed by the European Social Fund (EFOP-3.6.3-VEKOP-16-2017-00002) and the Hungarian National Excellence Grant 2018-1.2.1-NKP-00008.

\newpage

\newpage

\clearpage

\bibliography{main}
\bibliographystyle{icml2020}

\end{document}